# Classification of Major Depressive Disorder via Multi-Site Weighted LASSO Model

Dajiang Zhu[1], Brandalyn C. Riedel[1], Neda Jahanshad[1], Nynke A. Groenewold[2,3],
Dan J. Stein[3], Ian H. Gotlib[4], Matthew D. Sacchet[5], Danai Dima[6,7], James H. Cole[8],
Cynthia H.Y. Fu[9], Henrik Walter[10], Ilya M. Veer[11], Thomas Frodl[11,12],
Lianne Schmaal[13,14,15], Dick J. Veltman[15], Paul M. Thompson[1]

[1]Imaging Genetics Center, USC Stevens Neuroimaging and Informatics Institute, Keck School of Medicine of the University of Southern California, CA, USA;
[2]BCN NeuroImaging Center and Department of Neuroscience of the University of Groningen, University Medical Center Groningen, The Netherlands;
[3]Dept of Psychiatry and Mental Health, University of Cape Town, South Africa;
[4]Neurosciences Program and Department of Psychology, Stanford University, CA, USA;
[5]Department of Psychiatry and Behavioral Sciences, Stanford University, CA, USA;
[6]Dept of Neuroimaging, Institute of Psychiatry, Psychology and Neuroscience, King's College London, UK;
[7]Dept of Psychology, School of Arts and Social Science, City, University of London, UK;
[8]Department of Medicine, Imperial College London, UK;
[9]Department of Psychological Medicine, King's College London, UK;
[10]Dept of Psychiatry and Psychotherapy, Charité Universitätsmedizin Berlin, Germany;
[11]Department of Psychiatry, Trinity College Dublin, Ireland;
[12]Dept of Psychiatry and Psychotherapy, Otto von Guericke University Magdeburg, Germany;
[13]Dept of Psychiatry and Neuroscience Campus Amsterdam, VU University Medical Center, The Netherlands;
[14]Orygen, The National Centre of Excellence in Youth Mental Health, Australia;
[15]Center for Youth Mental Health, The University of Melbourne, Australia

**Abstract.** Large-scale collaborative analysis of brain imaging data, in psychiatry and neurology, offers a new source of statistical power to discover features that boost accuracy in disease classification, differential diagnosis, and outcome prediction. However, due to data privacy regulations or limited accessibility to large datasets across the world, it is challenging to efficiently integrate distributed information. Here we propose a novel classification framework through multi-site weighted LASSO: each site performs an iterative weighted LASSO for feature selection separately. Within each iteration, the classification result and the selected features are collected to update the weighting parameters for each feature. This new weight is used to guide the LASSO process at the next iteration. Only the features that help to improve the classification accuracy are preserved. In tests on data from five sites (299 patients with major depressive disorder (MDD) and 258 normal controls), our method boosted classification accuracy for MDD by 4.9% on average. This result shows the potential of the proposed new strategy as an effective and practical collaborative platform for machine learning on large scale distributed imaging and biobank data.

**Keywords:** MDD, weighted LASSO

## 1 Introduction

Major depressive disorder (MDD) affects over 350 million people worldwide [1] and takes an immense personal toll on patients and their families, placing a vast economic burden on society. MDD involves a wide spectrum of symptoms, varying risk factors, and varying response to treatment [2]. Unfortunately, early diagnosis of MDD is challenging and is based on behavioral criteria; consistent structural and functional brain abnormalities in MDD are just beginning to be understood. Neuroimaging of large cohorts can identify characteristic correlates of depression, and may also help to detect modulatory effects of interventions, and environmental and genetic risk factors. Recent advances in brain imaging, such as magnetic resonance imaging (MRI) and its variants, allow researchers to investigate brain abnormalities and identify statistical factors that influence them, and how they relate to diagnosis and outcomes [12]. Researchers have reported brain structural and functional alterations in MDD using different modalities of MRI. Recently, the ENIGMA-MDD Working Group found that adults with MDD have thinner cortical gray matter in the orbitofrontal cortices, insula, anterior/posterior cingulate and temporal lobes compared to healthy adults without a diagnosis of MDD [3]. A subcortical study – the largest to date – showed that MDD patients tend to have smaller hippocampal volumes than controls [4]. Diffusion tensor imaging (DTI) [5] reveals, on average, lower fractional anisotropy in the frontal lobe and right occipital lobe of MDD patients. MDD patients may also show aberrant functional connectivity in the default mode network (DMN) and other task-related functional brain networks [6].

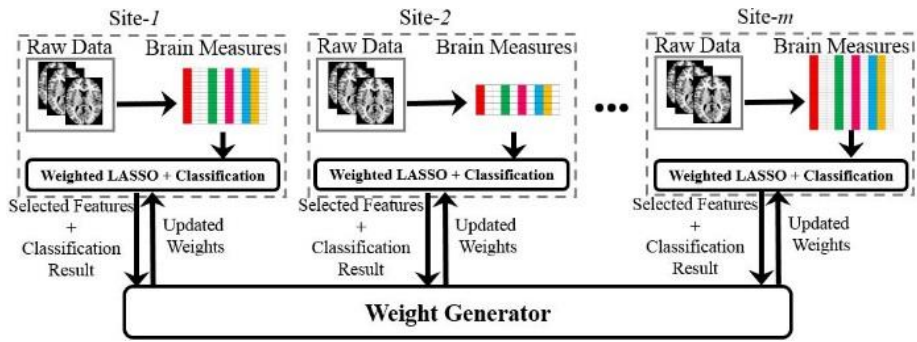

**Fig. 1.** Overview of our proposed framework.

Even so, classification of MDD is still challenging. There are three major barriers: first, though significant differences have been found, these previously identified brain regions or brain measures are not always consistent markers for MDD classification [7]; second, besides T1 imaging, other modalities including DTI and functional magnetic resonance imaging (fMRI) are not commonly acquired in a clinical setting; last, it is not always easy for collaborating medical centers to perform an integrated data analysis due to data privacy regulations that limit the exchange of individual raw data and due to large transfer times and storage requirements for thousands of images. As biobanks grow, we need an efficient platform to integrate predictive information from multiple centers; as the available datasets increase, this effort should increase the statistical power to identify predictors of disease diagnosis and future outcomes, beyond what each site could identify on its own.

In this study, we introduce a multi-site weighted LASSO (MSW-LASSO) model to boost classification performance for each individual participating site, by integrating their knowledge for feature selection and results from classification. As shown in **Fig. 1**, our proposed framework features the following characteristics: (1) each site retains their own data and performs weighted LASSO regression, for feature selection, locally; (2) only the selected brain measures and the classification results are shared to other sites; (3) information on the selected brain measures and the corresponding classification results are integrated to generate a unified weight vector across features; this is then sent to each site. This weight vector will be applied to the weighted LASSO in the next iteration; (4) if the new weight vector leads to a new set of brain measures and better classification performance, the new set of brain measures will be sent to other sites. Otherwise, it is discarded and the old one is recovered.

## 2 Methods

### 2.1 Data and demographics

For this study, we used data from five sites across the world. The total number of participants is 557; all of them were older than 21 years old. Demographic information for each site's participants is summarized in **Table 1**.

|   | Sites | Total N | Total N of MDD patients (%) | Total N of Controls (%) | Age of Controls (Mean ± SD; y) | Age of MDD (Mean ± SD; y) | % Female MDD | % Female Total |
|---|---|---|---|---|---|---|---|---|
| 1 | Groningen | 45 | 22 (48.89%) | 23 (51.11%) | 42.78 ± 14.36 | 43.14 ± 13.8 | 72.73 | 73.33 |
| 2 | Stanford | 110 | 54 (49.09%) | 56 (50.91) | 38.17 ± 9.97 | 37.75 ± 9.78 | 57.41 | 60.00 |
| 3 | BRCDECC | 130 | 69 (53.08%) | 61 (46.92%) | 51.72 ± 7.94 | 47.85 ± 8.91 | 68.12 | 60.77 |
| 4 | Berlin | 172 | 101 (58.72%) | 71 (41.28%) | 41.09 ± 12.85 | 41.21 ± 11.82 | 64.36 | 60.47 |
| 5 | Dublin | 100 | 53 (53%) | 47 (47%) | 38.49 ± 12.37 | 41.81 ± 10.76 | 62.26 | 57.00 |
|   | Combined | 557 | 299 (53.68%) | 258 (46.32$) |   |   |   |   |

**Table 1.** Demographics for the five sites participating in the current study.

### 2.2 Data preprocessing

As in most common clinical settings, only T1-weighted MRI brain scans were acquired at each site; quality control and analyses were performed locally. Sixty-eight (34 left/34 right) cortical gray matter regions, 7 subcortical gray matter regions and the lateral ventricles were segmented with FreeSurfer [8]. Detailed image acquisition, pre-processing, brain segmentation and quality control methods may be found in [3, 9]. Brain measures include cortical thickness and surface area for cortical regions and volume for subcortical regions and lateral ventricles. In total, 152 brain measures were considered in this study.

### 2.3 Algorithm overview

To better illustrate the algorithms, we define the following notations:

1. $F_i$: The selected brain measures (features) of *Site-i*;
2. $A_i$: The classification performance of *Site-i*;
3. *W*: The weight vector;
4. *w-LASSO (W, $D_i$)*: Performing weighted LASSO on $D_i$ with weight vector – W;
5. *SVM ($F_i$, $D_i$)*: Performing SVM classifier on $D_i$ using the feature set - $F_i$;

The algorithms have two parts that are run at each site, and an integration server. At first, the integration server initializes a weight vector with all ones and sends it to all sites. Each site use this weight vector to conduct weighted LASSO (**Section 2.6**) with their own data locally. If the selected features have better classification performance, it will send the new features and the corresponding classification result to the integration server. If there is no improvement in classification accuracy, it will send the old ones. After the integration server receives the updates from all sites, it generates a new weight vector (**Section 2.5**) according to different feature sets and their classification performance. The detailed strategy is discussed in **Section 2.5**.

| *Algorithm 1 (Integration Server)* |
|---|
| 1. Initialize *W* (*with all features weighted as one*) |
| 2. Send *W* to **all sites** |
| 3. while at least one site has improvement on *A* |
| 4.     update *W* (**Section 2.5**) |
| 5.     Send *W* to **all sites** |
| 6. end while |
| 7. Send *W* with null to **all sites** |

**Table 2.** Main steps of Algorithm 1.

| *Algorithm 2 (Site-i)* |
|---|
| 1. $F_i \leftarrow \emptyset, A_i \leftarrow 0$ |
| 2. while received *W* is not null |
| 3.     $F_i' \leftarrow$ *w-LASSO (W, $D_i$)* (**Section 2.6**) |
| 4.     if $F_i' \neq F_i$ |
| 5.         $A_i' \leftarrow$ *SVM ($F_i'$, $D_i$)* |
| 6.         if $A_i' > A_i$ |
| 7.             send $F_i'$ and $A_i'$ to *Integration Server* |
| 8.             $F_i \leftarrow F_i', A_i \leftarrow A_i'$ |
| 9.         else send $F_i$ and $A_i$ to *Integration Server* |
| 10.         end if |
| 11.     end if |
| 12. end while |

**Table 3.** Main steps of Algorithm 2.

### 2.4 Ordinary LASSO and weighted LASSO

LASSO [11] is a shrinkage method for linear regression. The ordinary LASSO is defined as:

$$\hat{\beta}(\text{LASSO}) = \arg\min \|y - \sum_{i=1}^{n} x_i \beta_i\|^2 + \lambda \sum_{i=1}^{n} |\beta_i| \qquad (1)$$

Y and x are the observations and predictors. λ is known as the sparsity parameter. It minimizes the sum of squared errors while penalizing the sum of the absolute values of the coefficients - β. As LASSO regression will force many coefficients to be zero, it is widely used for variable selection [11].

However, the classical LASSO shrinkage procedure might be biased when estimating large coefficients [12]. To alleviate this risk, adaptive LASSO [12] was developed and it tends to assign each predictor with different penalty parameters. Thus it can avoid having larger coefficients penalized more heavily than small coefficients. Similarly, the motivation of multi-site weighted LASSO (MSW-LASSO) is to penalize different predictors (brain measures), by assigning different weights, according to its classification performance across all sites. Generating the weights for each brain measure (feature) and the MSW-LASSO model are discussed in **Section 2.5** and **2.6**.

### 2.5 Generation of a Multi-Site Weight

In **Algorithm 1**, after the integration server receives the information on selected features (brain measures) and the corresponding classification performance of each site, it generates a new weight for each feature. The new weight for the $f^{th}$ feature is:

$$W_f = \sum_{s=1}^{m} \Psi_{s,f} \, A_s \, P_s / m \qquad (2)$$

$$\Psi_{s,f} = \begin{cases} 1, if & \text{the } f^{th} \text{ feature was selected in site} - s \\ & 0, otherwise \end{cases} \qquad (3)$$

Here $m$ is the number of sites. $A_s$ is the classification accuracy of site-$s$. $P_s$ is the proportion of participants in site-$s$ relative to the total number of participants at all sites. Eq. (3) penalizes the features that only "survived" in a small number of sites. On the contrary, if a specific feature was selected by all sites, meaning all sites agree that this feature is important, it tends to have a larger weight. In Eq. (2) we consider both the classification performance and the proportion of samples. If a site has achieved very high classification accuracy and it has a relatively small sample size compared to other sites, the features selected will be conservatively "recommended" to other sites. In general, if the feature was selected by more sites and resulted in higher classification accuracy, it has larger weights.

### 2.6 Multi-Site weight LASSO

In this section, we define the multi-site weighted LASSO (MSW-LASSO) model:

$$\hat{\beta}_{MSW-Lasso} = \arg\min \|y - \sum_{i=1}^{n} x_i \beta_i\|^2 + \lambda \sum_{i=1}^{n} (1 - \sum_{s=1}^{m} \Psi_{s,i} \, A_s \, P_s / m) |\beta_i| \qquad (4)$$

Here $x_i$ represents the MRI measures after controlling the effects of age, sex and intracranial volume (ICV), which are managed within different sites. $y$ is the label indicating MDD patient or control. $n$ is the 152 brain measures (features) in this study. In our MSW-LASSO model, a feature with larger weights implies higher classification performance and/or recognition by multiple sites. Hence it will be penalized less and has a greater chance of being selected by the sites that did not consider this feature in the previous iteration.

## 3 Results

### 3.1 Classification improvements through the MSW-LASSO model

In this study, we applied **Algorithm 1** and **Algorithm 2** on data from five sites across the world. In the first iteration, the integration server initialized a weight vector with all ones and sent it to all sites. Therefore, these five sites conducted regular LASSO regression in the first round. After a small set of features was selected using similar strategy in [9] within each site, they performed classification locally using a support vector machine (SVM) and shared the best classification accuracy to the integration server, as well as the set of selected features. Then the integration server generated the new weight according to Eq. (2) and sent it back to all sites. From the second iteration, each site performed MSW-LASSO until none of them has improvement on the classification result. In total, these five sites ran MSW-LASSO for six iterations; the classification performance for each round is summarized in Fig. 2 (a-e).

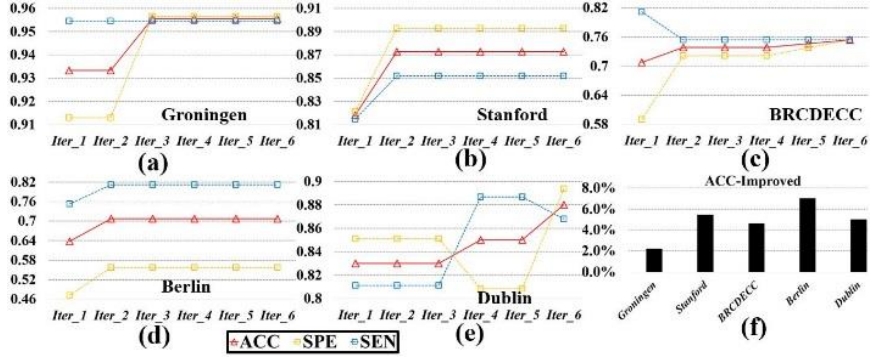

**Fig. 2.** Applying MSW-LASSO to the data coming from five sites (a-e). Each subfigure shows the classification accuracy (ACC), specificity (SPE) and sensitivity (SEN) at each iteration. (f) shows the improvement in classification accuracy at each site after performing MSW-LASSO.

Though the Stanford and Berlin sites did not show any improvements after the second iteration, the classification performance at the BRCDECC site and Dublin continued improving until the sixth iteration. Hence our MSW-LASSO terminated at the sixth round. **Fig. 2f** shows the improvements of classification accuracy for all five sites - the average improvement is 4.9%. The sparsity level of the LASSO is set as 16% - which means that 16% of 152 features tend to be selected in the LASSO process. **Section 3.3** shows the reproducibility of results with different sparsity levels. When conducing SVM classification, the same kernel (RBF) was used, and we performed a grid search for possible parameters. Only the best classification results are adopted.

### 3.2 Analysis of MSW-LASSO features

In the process of MSW-LASSO, only the new set of features resulting in improvements in classification are accepted. Otherwise, the prior set of features is preserved. The new features are also "recommended" to other sites by increasing the correspond-ing weights of the new features. **Fig. 3** displays the changes of the involved features through six iterations and the top 5 features selected by the majority of sites.

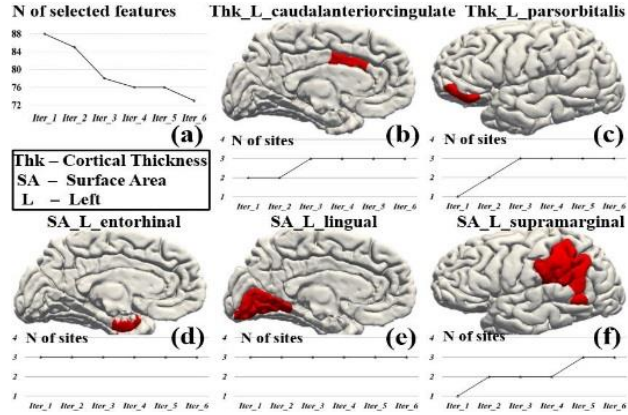

**Fig. 3.** (a) Number of involved features through six iterations. (b-f) The top five consistently selected features across sites. Within each subfigure, the top showed the locations of the corresponding features and the bottom indicated how many sites selected this feature through the MSW-LASSO process. (b-c) are cortical thickness and (d-f) are surface area measures.

At the first iteration, there are 88 features selected by five sites. This number decreases over MSW-LASSO iterations. Only 73 features are preserved after six iterations but the average classification accuracy increased by 4.9%. Moreover, if a feature is originally selected by the majority of sites, it tends to be continually selected after multiple iterations (**Fig. 3d-e**). For those "promising" features that are accepted by fewer sites at first, they might be incorporated by more sites as the iteration increased (**Fig. 2b-c, f**).

### 3.3   Reproducibility of the MSW-LASSO

| Selected Features | Improvement, in % | | | Selected features | Improvement, in % | | |
|---|---|---|---|---|---|---|---|
| | ACC | SPE | SEN | | ACC | SPE | SEN |
| **13%** | 3.1 | 1.8 | 4.4 | **33%** | 2.6 | 3.1 | 2.5 |
| **20%** | 3.9 | 1.4 | 6.0 | **36%** | 1.7 | 2.1 | 1.5 |
| **23%** | 3.8 | 2.9 | 4.4 | **40%** | 2.5 | 4.1 | 1.4 |
| **26%** | 4.3 | 3.4 | 5.2 | **43%** | 3.1 | 1.1 | 5.0 |
| **30%** | 2.9 | 3.0 | 2.9 | **46%** | 2.8 | 3.9 | 1.9 |

**Table 4.** Reproducibility results with different sparsity levels. The column of selected features represents the percentage of features preserved during the LASSO procedure, and the average improvement in accuracy, sensitivity, and specificity by sparsity.

For LASSO-related problems, there is no closed-form solution for the selection of sparsity level; this is highly data dependent. To validate our MSW-LASSO model, we repeated **Algorithm 1** and **Algorithm 2** at different sparsity levels, which leads to preservation of different proportions of the features. The reproducibility performance of our proposed MSW-LASSO is summarized in **Table 4**.

# 4 Conclusion and Discussion

Here we proposed a novel multi-site weighted LASSO model to heuristically improve classification performance for multiple sites. By sharing the knowledge of features that might help to improve classification accuracy with other sites, each site has multiple opportunities to reconsider its own set of selected features and strive to increase the accuracy at each iteration. In this study, the average improvement in classification accuracy is 4.9% for five sites. We offer a proof of concept for distributed machine learning that may be scaled up to other disorders, modalities, and feature sets.

# 5 References

1. World Health Organization. World Health Organization Depression Fact sheet, No. 369. (2012). Available from: http://www.who.int/mediacentre/factsheets/fs369/en/.
2. Fried, E.I., et al. "Depression is more than the sum score of its parts: individual DSM symptoms have different risk factors." **Psych Med.** 2067-2076 (2014).
3. Schmaal, L., et al. "Cortical abnormalities in adults and adolescents with major depression based on brain scans from 20 cohorts worldwide in the ENIGMA Major Depressive Disorder Working Group." **Mol Psych.** doi: 10.1038/mp.2016.60 (2016).
4. Schmaal, L., et al. "Subcortical brain alterations in major depressive disorder: findings from the ENIGMA Major Depressive Disorder working group." **Mol Psych** 806-812 (2016).
5. Liao, Y., et al. "Is depression a disconnection syndrome? Meta-analysis of diffusion tensor imaging studies in patients with MDD." **J Psych & Neurosci**. 49 (2013).
6. Sambataro, F., et al. "Revisiting default mode network function in major depression: evidence for disrupted subsystem connectivity." **Psychl Med**. 2041-2051 (2014).
7. Lo, A., et al. "Why significant variables aren't automatically good predictors." **PNAS**. 13892-13897 (2015).
8. https://surfer.nmr.mgh.harvard.edu/
9. Zhu, D., et al. Large-scale classification of major depressive disorder via distributed Lasso. **Proc. of SPIE**, 10160 (2017).
10. Tibshirani, R., "Regression shrinkage and selection via the LASSO." **Journal of the Royal Statistical Society**. 58: 267–288 (1996).
11. Li, Qingyang, et al., "Large-Scale Collaborative Imaging Genetics Studies of Risk Genetic Factors for Alzheimer's Disease Across Multiple Institutions." **MICCAI**. 335-343 (2016).
12. Zou, H., "The adaptive LASSO and its oracle properties." **J. Amer. Statist. Assoc** 101(476):1418-1429 (2006).
13. Koutsouleris, N., et al. Individualized differential diagnosis of schizophrenia and mood disorders using neuroanatomical biomarkers. **Brain**, *138*(7), 2059-2073 (2015).

*

* Supported in part by NIH grant U54 EB020403; see ref. 3 for additional support to co-authors for cohort recruitment.